\pdfoutput=1

\documentclass[11pt]{article}

\usepackage{acl}

\usepackage{times}
\usepackage{latexsym}
\usepackage[T1]{fontenc}
\usepackage[utf8]{inputenc}
\usepackage{microtype}
\usepackage{inconsolata}

\usepackage{booktabs}
\usepackage{float}
\usepackage{parskip}
\usepackage{color}
\usepackage{multirow}
\usepackage{enumitem}
\usepackage{graphicx}
\usepackage{inconsolata}
\usepackage{tipa}
\usepackage{wasysym}
\usepackage[skins,breakable]{tcolorbox}
\usepackage{linguex}

\definecolor{burntorange}{HTML}{bf5700}
\newtcolorbox[list inside=prompt,auto counter,number within=section]{prompt}[1][]{
    colbacktitle=burntorange,
    colframe=burntorange,
    fontupper=\footnotesize,
    boxsep=5pt,
    left=0pt,
    right=0pt,
    top=0pt,
    bottom=0pt,
    boxrule=1pt,
    enhanced, 
    breakable,
    skin first=enhanced,
    skin middle=enhanced,
    skin last=enhanced,
    #1,
}

\setlist{nosep}

\title{Do \emph{they} mean `us'?\\ Interpreting Referring Expression variation under Intergroup Bias}

\author{
    Venkata S\ Govindarajan\textsuperscript{\textbullseye}\qquad Matianyu Zang\textsuperscript{\sun}\\\textbf{Kyle Mahowald}\textsuperscript{\textramshorns}\qquad \textbf{David I. Beaver}\textsuperscript{\textramshorns}\qquad \textbf{Junyi Jessy Li}\textsuperscript{\textramshorns}\\
    \textsuperscript{\textbullseye}Department of Computer Science, Ithaca College\\
    \textsuperscript{\sun}Department of Computer Science, Brown University\\
    \textsuperscript{\textramshorns}Department of Linguistics, The University of Texas at Austin\\
    \textsuperscript{\textbullseye}\texttt{vgovindarajan@ithaca.edu}\qquad\textsuperscript{\sun}\texttt{zangmatianyu@gmail.com}
  }

\begin{document}

\setlength{\Exlabelwidth}{1em}
\setlength{\Exlabelsep}{1em}
\setlength{\SubExleftmargin}{1em}
\setlength{\Extopsep}{0.5\baselineskip}

\maketitle

\begin{abstract}
The variations between in-group and out-group speech (intergroup bias) are subtle and could underlie many social phenomena like stereotype perpetuation and implicit bias. In this paper, we model intergroup bias as a \emph{tagging task} on English sports comments from forums dedicated to fandom for NFL teams. We curate a dataset of over 6 million game-time comments from opposing perspectives (the teams in the game), each comment grounded in a \emph{non-linguistic description of the events that precipitated these comments} (live win probabilities for each team). Expert and crowd annotations justify modeling the bias through tagging of implicit and explicit referring expressions and reveal the rich, contextual understanding of language and the world required for this task. For large-scale analysis of intergroup variation, we use LLMs for automated tagging, and discover that LLMs occasionally perform better when prompted with \emph{linguistic descriptions} of the win probability at the time of the comment, rather than numerical probability. Further, large-scale tagging of comments using LLMs uncovers \textbf{linear variations in the form of referent across win probabilities} that distinguish  in-group and out-group utterances.
\end{abstract}

\section{Introduction}
\label{sec:intro}
\begin{figure}[t]
    \centering
    \includegraphics[width=\linewidth]{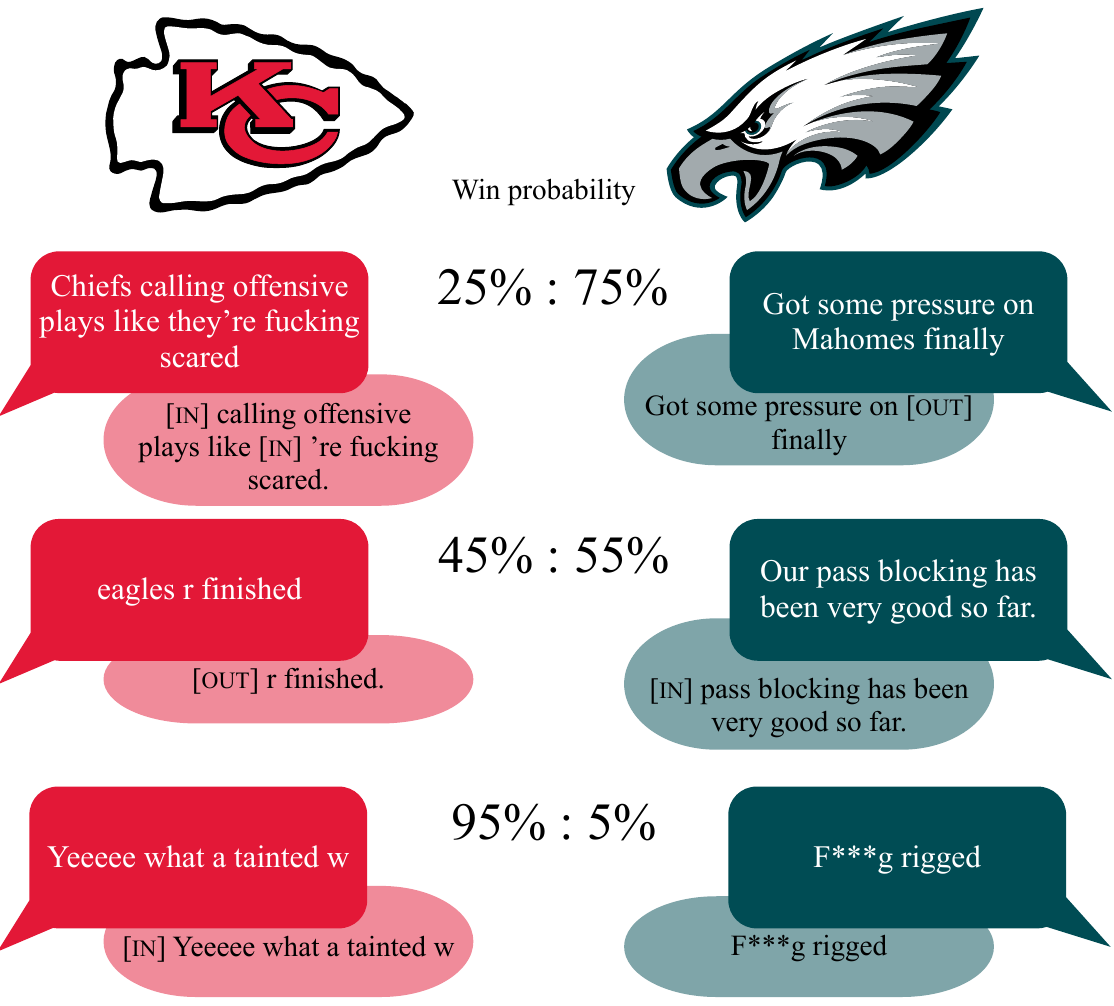}
    \caption{We construct a parallel language corpus of comments from NFL team subreddits, grounding each comment in the live win probabilities. We then tag relevant entities in each comment with intergroup tags using annotators and train LLMs to tag intergroup references.}
    \label{fig:flowchart}
    \vspace{-\baselineskip}
\end{figure}

Social bias in language is generally studied by identifying undesirable language use towards a specific demographic group~\citep{kaneko-bollegala-2019-gender, sheng-etal-2019-woman, sap-etal-2020-social, webson-etal-2020-undocumented, Pryzant2020, sheng-etal-2020-towards}; However, we can enrich our understanding of bias in communication by understanding it as differences in behavior situated in social relationships. Intergroup bias is the social bias stemming from the intergroup relationship between the speaker and target reference of an utterance.~\citep{maass_language_1989, maass_linguistic_1999}. \citet{govindarajan-etal-2023-people} modeled intergroup relationships (in-group and out-group) and interpersonal emotions in interpersonal English language tweets at the utterance level, finding systematic interactions between these two parameters. While neural models based on LLMs can be trained to discriminate in-group and out-group utterances, causal probing of these models was inconclusive~\citep{govindarajan-etal-2023-counterfactual}. Crucially, these studies still left unanswered a major question with respect to how the intergroup bias operates in the real world:

\begin{quote}
    How does language systematically change \emph{with respect to state-of-the-world} when referring to an individual in one's in-group versus out-group? 
\end{quote}

In this work, we take major steps towards answering this question, using a task architecture in the classical NLP pipeline~\citep{manning-etal-2014-stanford}. Earlier work in the Linguistic Intergroup Bias (LIB) hypothesis~\citep{maass_linguistic_1999} focused exclusively on the predicate, and the bias was described using an ad-hoc lexical categorization system~\citep{semin_cognitive_1988} over predicates. However, the \textbf{form of referencing} the in-group or out-group can reveal subtle biases as well. Consider the comments in Figure~\ref{fig:flowchart}, sampled from two fan forums discussing a live NFL game and part of the dataset described in this paper. Commenters refer to their in-group (team they support) and out-group (opponent team in that game) by name, sub-groups, pronouns as well as implicitly --- sometimes they choose not to refer to either group at all. By \emph{tagging} these references with appropriate labels which denote the relationship of the referent to the speaker, we can ask: How does the intergroup bias manifest in referent forms?

To answer this question, we introduce a new dataset of interpersonal language --- comments from game threads on online forums dedicated to fandoms for teams in the National Football League (NFL). Through careful data curation, we construct a parallel corpus of sports comments, with comments from fans of both teams in a game, \emph{aligned} in time and \emph{grounded} in win probabilities (WP). By focusing on referring expressions, we can formulate investigating the intergroup bias as a \textbf{tagging} task: given a comment, the group affiliation of the writer, and the state-of-the-world, return a \emph{tagged comment} with appropriate referring expressions tagged as {\scshape[in]}, {\scshape[out]} or {\scshape[other]} (see Figure~\ref{fig:flowchart}). Annotation and preliminary analysis reveal that the form of the referent that speakers use when referring may have systematic intergroup variations.

We train Large Language Models (LLMs) to automate tagging on our dataset, and examine their performance on our task. Few-shot performance on GPT-4o improves when using linguistic descriptions of WPs; fine-tuned Llama-3 models performed better, although incorporating WP had little effect. Using our best performing model to tag 100,000 comments from our dataset, we discover two striking linguistic behaviors at scale:

\begin{enumerate}
    \item Higher the win probability for the \emph{in-group}, the more likely commenters are to \textbf{abstract away} from referring to the in-group. This trend is remarkably linear across win probabilities for all types of in-group references.
    \item References to out-groups by commenters are rarer than in-group references, and remain stable over all win probabilities for the in-group.
\end{enumerate}

These findings add much needed color to the LIB hypothesis --- natural language is productive, and commenters can express their (implicit) intergroup bias in different ways. This work also lays the foundation for future explorations of other intergroup variations (in event descriptions, for example) in sports-talk and other domains. We share all our code and data online\footnote{ \url{https://github.com/venkatasg/intergroup-nfl}}.

\section{Background and Related Work}
\label{sec:background}
\paragraph{Intergroup bias} Linguistic Intergroup Bias  (LIB) theory~\cite{maass_language_1989,maass_linguistic_1999} hypothesizes that stereotypes are transmitted and persist in communication through systematic linguistic asymmetry ---  socially desirable in-group behaviors and socially undesirable out-group behaviors are encoded at a higher level of abstraction. The LIB has been reproduced in psychological experiments and analyses~\cite{Anolli2006LinguisticIB, gorham_news_2006}; it has also been used as an indicator for a speaker’s prejudicial attitudes~\cite{hippel_linguistic_1997}, and racism~\cite{schnake_modern_1998}.

~\citet{govindarajan-etal-2023-people, govindarajan-etal-2023-counterfactual} take  inspiration from the LIB at large to study intergroup bias as a general phenomenon in online language use. While they find regularities in its variation with emotion that neural models can `learn' to identify in-group and out-group utterances more accurately than humans, probing experiments fail to describe human-observable intergroup variations in language.  This work studies a much larger dataset than in their work, and by modeling the bias as a tagging task to referents, we discover characteristic lexical variations at scale that complement LIB findings. 

\paragraph{Sports language} Language use in the domain of sports has been a rich source of analyses and studies within computational linguistics, including from the perspective of quantifying \emph{social biases}. \citet{merullo-etal-2019-investigating} studied commentator racial biases in descriptions of football players, reaffirming previous findings illustrating clear differences in terms of sentiment descriptions (white players were more likely to be described as intelligent), and name itself (white players were more likely to be referred to by their first name). \citet{basketball} focused on one aspect of language usage among (and between) fans of NBA teams: intra-group behavior with and without social contact with the out-group. They find that fans with intergroup contact are more likely to use negative language --- they were more polarized than before.

Our work differs from previous work in two major ways. Firstly, we focus on the intergroup bias --- how do fans talk about their team (in-group), versus the opponent (out-group)? Secondly, this paper \textbf{grounds} the analysis of intergroup bias in numerical descriptions of the state-of-the-world. The state-of-the-world in a sports game at any moment can be described using the scoreboard, thus providing grounding for utterances follow. Non-linguistic, numerical descriptions of the events that precipitate an utterance overcome the drawbacks of using ad-hoc, proximally derived metrics like \emph{social desirability} (in LIB) or \emph{affect} (in \citet{govindarajan-etal-2023-counterfactual}) as an axis to study linguistic variation. As we shall describe in \textsection\ref{sec:data}, sports games, and in particular NFL games, are rich with statistical information amenable to describe the state-of-the-world on a well-calibrated numerical scale.

\section{Data \& Annotation}
\label{sec:data}
\subsection{Dataset}
 
\paragraph{Data \& Preprocessing} Our new dataset of intergroup language comes from Reddit --- specifically subreddits dedicated to fandoms for each of the 32 teams in the NFL. During the NFL season, each subreddit has \emph{game threads} --- posts created by moderators on which fans can comment in tandem with the live game involving their team. Crucially, since every subreddit has their own thread, we effectively have a \textbf{parallel intergroup language dataset}; two teams and their fans commenting on \emph{the same game events}. Further, these subreddits are dedicated to individual team fandoms, so we can fairly assume that the team the subreddit represents is the in-group for all commenters.\footnote{Note that we focus on language of online commenters (fans) on Reddit, not \emph{commentators} for the game.}

We focus on all completed games from the 2021--22 and 2022--23 NFL seasons, and attempted to scrape all comments from the game threads for both teams involved in every game. Within comments from game threads, we filtered it down to comments that happened during active game-time, and removed comments that were only URL links. Overall, our raw data has over 6 million comments from 1104 game threads on 32 subreddits, grounded in 569 NFL games. 

\paragraph{Grounding football comments} American Football has some attractive features as a sport considering that our interest is in the \emph{language surrounding the events} in a game --- it is highly strategic, and outcomes are heavily dependent on a coach's strategies and plays in a (relatively) small number of discrete events~\citep[called \emph{plays,}][]{pelechrinis2016anatomy}. The state-of-the-world at any moment in a football game is determined by a variety of factors --- seconds remaining in half (and game), yard line, score differential, down, yards to go, home advantage, timeouts remaining, betting odds lines from Vegas, and so many more~\citep{horowitz2017nflscrapr, Yurko2018nflWARAR}. \citet{baldwin2021nflfastr} modeled the \textbf{Win Probability} (henceforth \textbf{WP}) of a team at any point during the game using a decision trees over the aforementioned features, building a well-calibrated model with low error. We chose WP as a succinct, non-linguistic description of the events preceding an utterance.

Using the \texttt{nflFastR}~\citep{baldwin2021nflfastr} package, we can obtain WPs for individual plays in each game, as well as the time of completion of a play. Combined with the timestamps at which the comments were submitted (obtained from the Reddit API), we build our parallel corpus of intergroup language grounded in win probabilities. The WP cleverly models the complexities of a real-world sporting event into one number that accurately models how \textbf{desirable} the state-of-the-world is to the in-group (see Figure~\ref{fig:flowchart}). 

\subsection{Tagging}
\label{subsec:tagging}

As we motivated in \textsection\ref{sec:intro} and Figure~\ref{fig:flowchart}, tagging references to entities enables us to perform analyses at scale \emph{and} discover individual lexical variations. Consider the following examples:

\ex. \label{ex:mult-football} \a. \label{ex:mult-football-a} Rams are gifting us a chance to win and we can’t take advantage. The f***!!!!
     \b. \label{ex:mult-football-b} if the ravens and chiefs beat these dudes by double digits then damn it so should we!
     
Even without contextual information about the game for the above comments, we see \emph{multiple} readily identifiable references to the in-group and out-group, within the same utterance. The words or phrases that refer to relevant individuals can now be tagged with in-group ({\scshape [in]}) or out-group ({\scshape [out]}) For instance, \ref{ex:mult-football} would be tagged thus:

\ex.\label{ex:tagged-exes} \a. {\scshape [out]} are gifting {\scshape [in]} a chance to win and {\scshape [in]} can’t take advantage. The f***!!!!
     \b. if {\scshape [other]} and {\scshape [other]} beat {\scshape [out]} by double digits then damn it so should {\scshape [in]}!

We define the in-group ({\scshape [in]}) as the team the commenter supports (and its fans), and the out-group ({\scshape [out]}) as the opponent in that particular game (and its fans). The spans `the ravens' and `chiefs' in \ref{ex:mult-football-b} are clearly not a reference to the in-group nor the opponent of the game. However, they are a reference to \emph{a group of interest in this domain} --- another NFL team and/or its fans. We consider these references to be {\scshape [other]}, and a special case of out-group references. 

Sometimes, the references to the in-group, out-group or other are not explicit. However, we can infer based on common-sense reasoning that the comment as whole, or a sentence in the comment, is \textbf{implicitly referring} to a relevant group:

\ex. What a conservative play call

There is no explicit word/phrasal reference to any team in the above comment. However, it is clear in context (the fan's team is losing, with WP of 9\%) that the commenter is referring to the in-group. To facilitate these implicit annotations, we sentence tokenize the comments in our dataset using Stanza~\citep{qi2020stanza}, append a sentence-level token {\scshape [sent]} before each sentence in every comment in our dataset. If the sentence as a whole is judged to implicitly refer to a relevant group, the {\scshape [sent]} token is replaced with the relevant tag.

\subsection{Annotation}

Annotators are presented with a comment from our dataset, the source subreddit (team) for the comment, the parent comment (if the comment is a reply in a thread), and the live score at the time of the comment. The task of tagging words and phrases from comments in our dataset with intergroup tags can be highly involved --- in addition to knowledge of American Football, commonsense reasoning over the meaning of an utterance in context of the live game, one needs knowledge of the teams and its players. For instance, in \ref{ex:wilson}, one needs to know that the commenter supports the Seahawks, and that there is a prominent player named Wilson, to accurately tag in context that Wilson indeed is an in-group reference.

\ex. \label{ex:wilson} Our oline should start holding since apparently it ’s okay now . Maybe Wilson can actually get some time to throw .

Implicit annotations on the {\scshape [sent]} token require a higher bar of reference, since all comments are about the game at hand and will involve both teams to some extent. For example, we judge the following comments to not have explicit or implicit references to any relevant groups of interest even though they are about the game:

\ex. \a. Fair enough !
     \b. winning cures all lmao
     \b. turning the game off , have a good day yall
     
In case it is impossible to verify an explicit or implicit reference, annotators are instructed to not highlight any parts of the comment. All annotators were free to search the web for names or expressions they were unfamiliar with, as well as refer to reports of the game to understand the utterance completely, and accurately tag all references. All annotation experiments were carried out using the \texttt{thresh.tools} annotation interface~\citep{heineman-etal-2023-thresh}. Annotators highlight spans within a comment and select from one of 3 tags, and select a confidence level from a five-point scale. 

\paragraph{Expert annotated dataset} We gather expert annotations for constructing a `gold' annotated dataset to evaluate crowd annotations and modeling moving forward. The first author of the paper annotated 1499 comments (randomly sampled from all game-time comments) for intergroup references based on a pre-defined, written protocol (described in detail in Appendix~\ref{app:ann}).  26.7\% of comments were judged to have no relevant intergroup reference, and in the remaining comments, references to the in-group (76.3\%) vastly out-number references to the out-group (14.6\%) or other. This is not surprising, since these are comments from forums dedicated to fandom of teams --- people are much more likely to talk about their team over the opponent. We partition our gold dataset into a test set of 318 datapoints, and a training set of 1181 datapoints. 

\paragraph{Crowd annotation} To understand our dataset further, we recruited three undergraduates to annotate the test split of our expert dataset. Our goals were to understand where disagreements arose, as well as how and when knowledge of the events in the game helped in disambiguating references. Annotators were given similar instructions, and were free to search the web and lookup statistics and reports on the game in question. We found in pilot experiments that the live-score was more interpretable to humans than WP, influencing our choice to provide that as context to annotators.

\section{Preliminary Analysis}
\label{sec:prelim}
\subsection{Annotation Analysis}
\label{subsubsec:annotation-analysis}

\paragraph{Inter-annotator agreement} Average Fleiss $\kappa$~\citep{fleiss} among crowd annotators is  0.69, indicating moderate agreement. In addition to the inter-annotator score, by counting exact-matches and weighting partial matches between individual crowd annotators and gold annotations, we calculate an `accuracy' score of $0.65\pm0.005$. This can be interpreted as a human ceiling for performance on this task, and characterizes its inherent subjectivity and difficulty.

\paragraph{Disagreement can be a signal} Looking at the source of disagreements among annotators (and between crowd annotators and experts) can give us insights into the nature of the task itself~\citep{atwell-etal-2022-role}, as well as why differences in judgements of intergroup affiliation can come down to annotator biases or judgement given context. For example, annotators disagree sometimes on what counts as a `reference':

\ex. \label{ex:disagree} \a.\label{ex:disagree-a}\textellipsis\textbf{Lambeau} has the second worst bathrooms .
    \b. \label{ex:disagree-b} Can’t do that against \textbf{an offense} this good.

\emph{Lambeau} in \ref{ex:disagree-a} was judged by the expert annotator to be a reference to the out-group in this context --- the opponent/out-group is the Green Bay Packers. However, to tag this referent {\scshape [out]}, annotators would need to know or deduce that Lambeau Field is the Green Bay Packers stadium, and judge that this constitutes a relevant intergroup reference. Thus, disambiguating some references can be time-consuming and hard. \emph{an offense} in \ref{ex:disagree-b} was judged by some annotators to refer to the out-group in context. However the generic nature of the referent lead other annotators to judge that this was an overall statement about the game, rather than an explicit reference. 

Whether or not examples in \ref{ex:disagree} contain references to the in/out-group is not simply a consequence of the difficulty of our task, or the inability for annotators to transparently describe the mental state of commenters. Rather, we need to analyze them as possibly another subtle influence of the intergroup bias itself --- demonstrated by questioning why commenters chose the forms in \ref{ex:disagree} rather than in \ref{ex:disagree-alt}, which convey the same meaning, and would be uncontroversial in annotation:

\ex. \label{ex:disagree-alt} \a. \textellipsis\textbf{the Packer's} stadium has the second worst bathrooms .
    \b. Can’t do that against \textbf{a Packers} offense\textellipsis

\subsection{Qualitative Analysis \& Trends}
\label{subsec:mereology}

\begin{figure}[t]
    \centering
    \includegraphics[width=\linewidth]{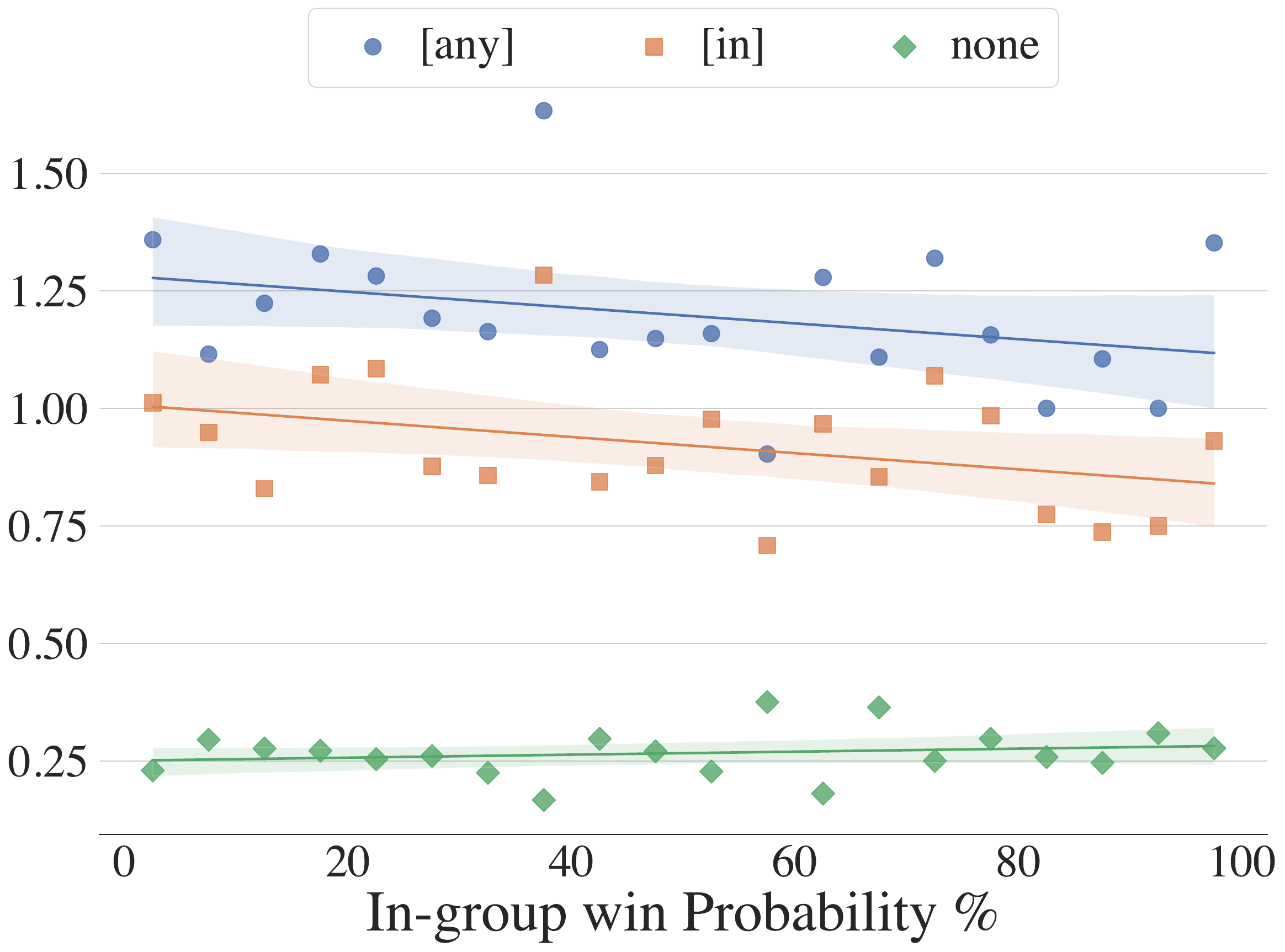}
    \caption{Per-comment frequency of in-group, any and `none' references in gold dataset over WP.}
    \label{fig:gold-trends}
\end{figure}

\paragraph{Mereology of referring expressions} Expert annotation revealed that commenters refer to groups of interest in a myriad of different ways. In the previous section, we liberally defined the annotation protocol for highlighting references to \emph{individuals} in the in-group, out-group and other. Using insights from mereology~\citep{sep-mereology}, we derive a taxonomy of parthood in intergroup relations, that defines what it means for a reference to constitute a reference towards the in-group/out-group/other:

\begin{enumerate}
    \item \textbf{People}: Names, nicknames, shirt numbers, initials, pronouns, etc. : \emph{Tua, TK87, he/him\textellipsis}
    \item \textbf{Subset of the team}: This refers to groups of players, or coachers, rather than just one player: \emph{the offense}, \emph{our defense}, \emph{o-line}, \textellipsis
    \item \textbf{Team}: Name of the team (\emph{rams, bills, cowboys}), nicknames (\emph{lambs, cowgirls}), city names(\emph{LA, Buffalo, Dallas}), pronominal expressions like \emph{our boys} for the in-group, pronouns like \emph{they/them} for the in-group and out-group, and many more.
    \item \textbf{Team plus supporters}: The first person pronouns \emph{we} and \emph{us}, but can also be done with the third person pronouns \emph{they} and \emph{them}. The latter of course, could also refer to out-group or other, and require context to disambiguate.
\end{enumerate}

The taxonomy above is ordered in order of increasing coverage of the whole group, by the referring part --- the size of the reference gets larger from people to the entire group. Thus, players are the smallest unit of reference within a group, and the team/organization plus its supporters constitute the largest possible reference to the group itself. 

\paragraph{Trends} The annotated dataset enables us to study qualitative trends, that will guide quantitative modeling analyses presented in \textsection\ref{sec:analysis}. We specifically focus on two phenomenon that are directly observable in the data and illustrated with examples --- diversity in form of referring expression, and trends over WP. Within the gold dataset, we can observe two clear trends by plotting the frequency of a feature of interest over comments that fall within a win probability (WP) window. Figure~\ref{fig:gold-trends} plots the frequency of any reference, in-group references, and `None' references over all 5\% WP windows:

\begin{enumerate}
    \item References to the in-group, and references to any group overall, go down with WP.
    \item `None' references increase steadily with WP.
\end{enumerate}

The steady increase in number of `None' references in higher WP windows is interesting, but requires robust analysis. While the trends observed in this section are not statistically significant, this can be attributed to the small sample size of only 1499 comments. The intergroup bias is a social phenomenon, and like many social phenomenon, we can make clear inferences at scale. Obtaining human annotated data at scale would be prohibitively hard and expensive in this setting --- we use LLMs, to automate this task, thus allowing us make inferences about trends in the intergroup bias as a function of WP. 

\section{Modeling intergroup bias with LLMs}
\label{sec:modeling}
Large Language Models (LLMs) have shown remarkable abilities in various domains over the last few years~\citep{srivastava2023beyond, Brown2020LanguageMA}. Our novel tagging framework to model intergroup bias requires linguistic understanding, knowledge of the NFL and its teams, as well as complex reasoning over why a commenter might choose certain word forms compatible with the state-of-the-world --- making LLMs well suited to this task. In this section, we design modeling experiments to tag comments from our dataset with intergroup labels with 2 goals:

\begin{itemize}
    \item Understand how LLMs statistically `reason' over meaning in context of an utterance and game state (WP) to tag comments.
    \item Our main objective is to discover hidden intergroup variations in referring expressions by tagging a large sample of comments from our raw, untagged data.
\end{itemize}

\begin{table*}[t]
    \centering
    \begin{tabular}{l|lrrrrrrr}
        \toprule
        \textbf{} & \textbf{Model} & \textbf{Random} & \textbf{Numeric} & \textbf{Numeric} & \textbf{No WP} & \textbf{No WP} & \textbf{Ling. WP} & \textbf{Ling WP} \\ 
        & \textbf{} & \textbf{Baseline} & \textbf{} & \textbf{WP+TS} & & \textbf{+TS} & & \textbf{+TS}\\\midrule
        \parbox[t]{2mm}{\multirow{3}{*}{\rotatebox[origin=c]{90}{\small~GPT-4o}}} & {\scshape[in]}  & 35.6(3.2) & 68.4(3.0) & 70.3(1.1) & 67.9(3.6) & 69.6(1.6) & 68.9(1.0) & \textbf{71.0}(1.0)\\
        & {\scshape[out]}  & 20.1(1.1) & \textbf{71.0}(1.8) & 68.2(2.0) & 63.8(5.3) & 65.5(3.1) & 65.5(4.1) & 68.6(2.8)\\
        & {\scshape[other]}  & 14.0(5.9) & 51.0(7.6) & 55.5(4.9) & 50.3(4.5) & 55.1(4.0) & 54.9(4.7) & \textbf{56.4}(4.1)\\
        & Overall  & 30.8(2.7) & 66.8(3.3) & 68.3(0.6) & 65.3(2.3) & 67.3(1.0) & 66.8(0.7) & \textbf{69.0}(1.1)\\\midrule
        \parbox[t]{2mm}{\multirow{3}{*}{\rotatebox[origin=c]{90}{\small~Llama-3-8b}}} & {\scshape[in]} & 35.6(3.2) & 74.2(0.5) & 73.8(0.3) & 73.9(0.6) & 73.8(1.2) & \textbf{74.3}(1.5) & 72.7(0.8)\\
        & {\scshape[out]}  & 20.1(1.1) & \textbf{62.5}(3.2) & 59.4(5.4) & 62.4(1.0) & 57.7(2.6) & 62.2(1.9) & 60.6(2.3)\\
        & {\scshape[other]}  & 14.0(5.9) & 61.0(1.5) & 61.2(7.5) & 60.9(7.1) & \textbf{63.2}(8.1) & 60.4(3.1) & 56.5(5.9)\\
        & Overall  & 30.8(2.7) & \textbf{71.0}(1.0) & 70.2(1.0) & 70.7(1.3) & 70.1(1.2) & 70.8(1.7) & 69.0(1.5)\\ \bottomrule
    \end{tabular}
    \caption{Mean F1 scores (S.D in parentheses) from few-shot experiments on GPT-4o and finetuning Llama-3-8b.}
    \label{tab:results}
\end{table*}

\subsection{Modeling conditions}

We focus on two specific models --- Llama-3-8B~\citep{llama3} and GPT-4o~\citep{gpt4o}. Both are decoder based models that perform best at a wide variety of benchmarks, and allow us to compare and contrast the performance of an open-weights model with finetuning, versus a larger closed model with few-shot prompting. Building upon previous work, we prompt both models with a combination of instructions, chain-of-thought explanations, and few-shot examples~\citep{NEURIPS2022_9d560961}. Llama-3 is prompted with the same input format, but we also finetune the model on the train split of our gold dataset. See Appendix~\ref{app:implement} for further details on training and inference.

\paragraph{CoT Explanations} We finetune Llama-3 with GPT-4o generated CoT explanations ~\citep{wadhwa-etal-2023-revisiting}. We first generate a  explanation from GPT-4o for each comment in our gold dataset using  instructions, few-shot examples, the target tagged comment and list of referring expressions provided as input to GPT-4o. All few-shot explanations were written by the first author, and examples were drawn from outside the gold dataset. 

Our task is framed end-to-end as the model receiving the untagged comment as input with some contextual information (in-group, out-group, WP, parent comment), and being asked to generate the comment with relevant words/phrases replaced with the appropriate tags. To understand the impact of WP on model performance we design 3 conditions

\begin{description}
    \item[Numeric WP] The model receives WP as a numeric input --- a percentage between 0 and 100 that is WP for the in-group.  
    \item[No WP] WP is not provided as input to the model, and the instructions nor few-shot explanations neither use nor mention it.
    \item[Linguistic WP] We experiment with providing WP as a scalar description of game state, from `Team A is very likely to win' to `Team B is very likely to win' based on the numeric WP corresponding to the comment. 
\end{description}

We also experimented with utilizing the WP to modify the temperature when decoding~\citep{atwell-etal-2022-role}. When temperature scaling (TS) is used, we set the temperature to $sin(\pi.WP)$ --- this pushes the LM to choose less likely words when the game's outcome is more uncertain. 

\paragraph{Evaluation} To evaluate the performance of a model on the test dataset, we report \textbf{micro-F1} scores for each of the three tags, and a weighted macro-F1 score overall. To give partial credit for the model's tagged output slightly overlapping with the gold tagged spans, we assign partial scores (0.5 and 0.25) for being within 3 and 5 characters of the correct tagged spans respectively.

\subsection{Results}
\label{subsec:football-results}

Table~\ref{tab:results} shows the results for both models in all conditions. We calculate and report the mean and standard deviation over 3 random seeds for each model under every condition. While both models exceed the human baseline performance that we calculated in \textsection\ref{sec:prelim}, Llama-3 nudges GPT-4o overall. By inspecting model generated outputs, we reason that GPT-4o performs better at identifying out-group references by names or nicknames due to its much larger size and more parametric knowledge.

\paragraph{WP helps\textellipsis~sometimes?} Including WP did not change the performance of Llama-3 noticeably. As we observed in annotation, there are few examples of comments being ambiguous enough that the state-of-the-world is enough to disambiguate what a reference could be. Entire classes of references (from our taxonomy in \textsection\ref{subsec:mereology}) are quite unambiguous even without whole-sentence context. 

We do observe however that providing WP in language form boosts the in-group and overall tagging performance of GPT-4o in few-shot settings, although this result is not statistically significant under a bootstrap test. Analysis of model's outputs reveal GPT-4o's fickleness and inability to reason over numerical scales --- for instance it reasons that WPs ranging from 1\% to as high as 41\% are `low' in its explanations. Further, it rarely uses the numbers to infer the WP for the out-group in explanations. Since we re-write low WPs with the name of the out-group (as winning) in the linguistic WP condition, this might explain the model's slight boost in performance.

While Llama-3's performance is better through finetuning, it does not benefit from incorporating WP in training or inference. We attempted scaling the loss during training with WP and expert annotator confidence ratings where available, but these didn't boost performance. Whether larger LLMs exhibit similar behaviors to GPT-4o when finetuned, we leave to future work.

\paragraph{Error analysis} While phrasing the WP with words improves tagging performance on in-group referents (over numerical WP as Table~\ref{tab:results} shows), especially with GPT-4o, performance on out-group references remains stable. However, we do observe the model making similar `errors' like annotators that we described in \textsection\ref{subsubsec:annotation-analysis} --- for instance, GPT-4o occasionally tagged \emph{a single WR} (an indefinite/generic entity) as out-group in \ref{ex:error-1a}, which was not judged to be a relevant referent in expert annotation. However, GPT-4o also makes some basic errors, such as tagging \emph{a catch} as out-group in \ref{ex:error-1b} as well.

\ex.\label{ex:error-1} \a.\label{ex:error-1a} \textellipsis~Did I hear that right? \textbf{They} don't have \textbf{a single WR} with a catch today??
     \b.\label{ex:error-1b} \textellipsis~Did I hear that right ? {\scshape [out]} don't have {\scshape [out]} with {\scshape [out]}??

\section{Analysis of model-tagged comments}
\label{sec:analysis}
\begin{figure}[t]
    \centering
    \includegraphics[width=\linewidth]{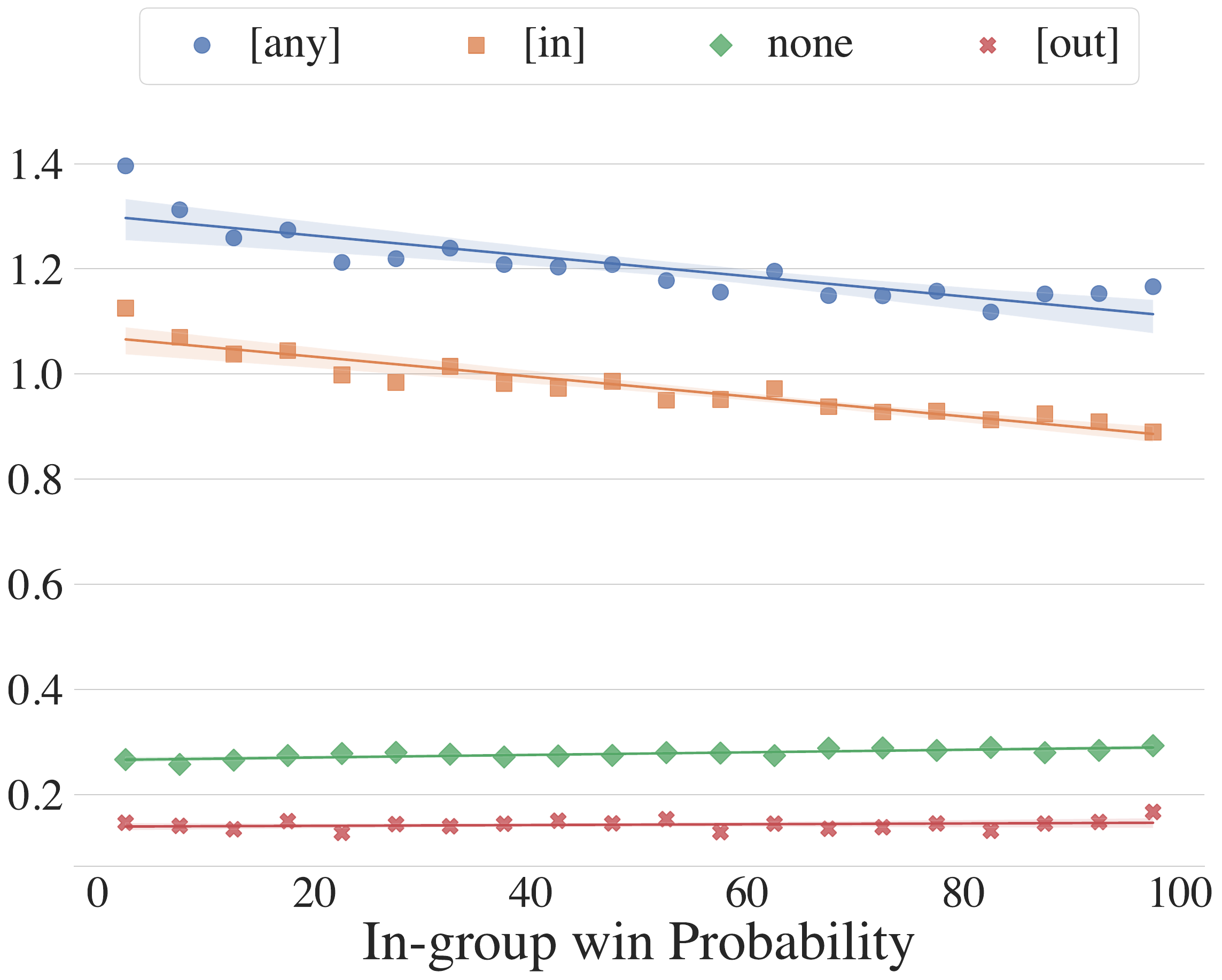}
    \caption{Per-comment frequency of various reference variables over all 5\% WP windows. A 95\% CI regression line is fit separately for each variable.}
    \label{fig:trends-1}
    \vspace{-\baselineskip}
\end{figure}

Our novel tagging framework is amenable to application on a large sample of our raw data, facilitating us to observe and analyze variations in how members refer to different groups as a function of WP. We sample 100,000 comments from a larger raw dataset, and apply our best performing finetuned Llama-3 model towards this task. Since WP could not be effectively incorporated to improve performance, we used the model finetuned with no WP information. Further, we verified there was no correlation between the model's accuracy and WP --- accuracy mostly followed comment density across WP (see Appendix~\ref{app:data}). After inference from our finetuned LLM, we use regular expressions to ensure that any obvious words (names and nicknames of teams, \emph{we/us}) were also tagged appropriately, and to count different referring expressions accounting for inflections.

\begin{figure}[t]
    \centering
    \includegraphics[width=\linewidth]{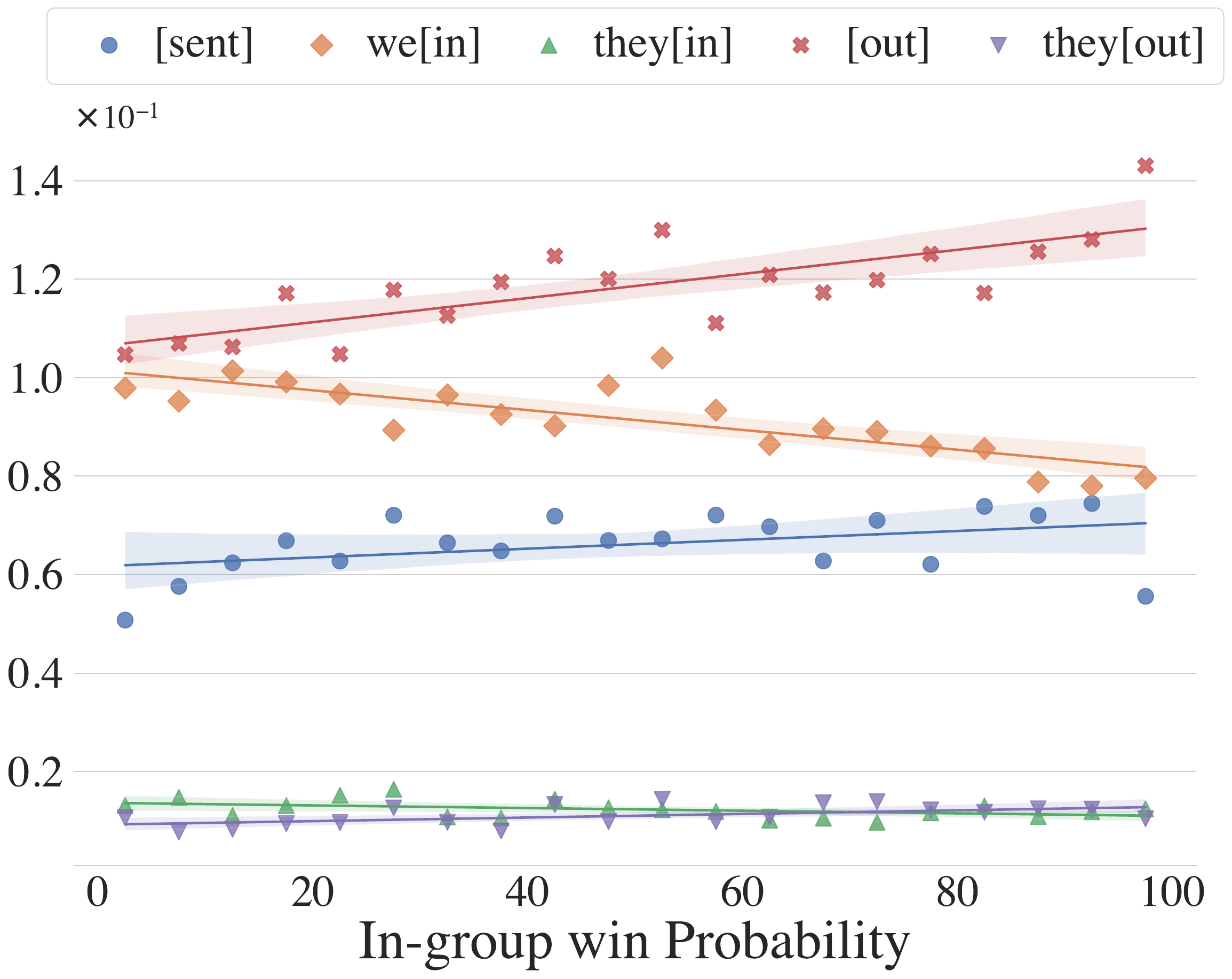}
    \caption{\emph{Normalized} per-comment frequency of various reference variables over all 5\% WP windows. A 95\% CI regression line is fit separately for each variable.}
    \label{fig:trends-2}
    \vspace{-\baselineskip}
\end{figure}

Figure~\ref{fig:trends-1} plots the frequency of different references over WP. We divide WP into 5\% windows, and count the number of comments that contain a specific tag (or tag-lexical item pair), and divide it by the number of comments within that window in the entire sample. Figure~\ref{fig:trends-2} plots a few more reference variables of interest and is similar except the variables are normalized by number of comments that contain \emph{any} reference. There are two findings we wish to highlight in these figures.
 
\paragraph{Winning trumps all} Figure~\ref{fig:trends-1} clearly shows a linear, decreasing tendency for commenters to refer to any entity (including the in-group) the more likely the team is to win. Observing a sample of high WP~\ref{ex:high-wp}) comments reveals an increased (positive) excitement and terseness with higher WPs:

\ex.\label{ex:high-wp} \a. HOLY S**T 
     \b. WHAT A THROW

Figure~\ref{fig:trends-2} illustrates that the tendency to refer less to the in-group is concurrent with an increased tendency to refer to the out-group, or to refer implicitly over a sentence (with {\scshape [sent]}) when referring at all, as WP increases. Overall, this paints a clear picture --- as the in-group is more likely  to win, commenters prefer to refer to the out-group or to refer to entities implicitly if they refer at all. As \ref{ex:high-wp} shows, they prefer to abstract away from talking about specific events in the game or referring to specific individuals, towards expressing excitement instead at high WPs.

\paragraph{WP as a well calibrated predictor} A striking feature of Figures~\ref{fig:trends-1} and \ref{fig:trends-2} is the linear relationships between reference variables and WP. Table~\ref{tab:slopes} estimates coefficients and R-squared for linear fits of various variables measured. We observe that with increasing WP, commenters are more likely to refer to the out-group with `they' than the in-group, which lends credence to our intuitions and research that shows \emph{they} being a classic `other-ing' term~\citep{riggins1997language}. Further, we observe that the estimated slope for in-group references ($-2.8e^-4$) is larger than the slope for references to the in-group using first person singular pronouns ($-2e-4$, labelled as \texttt{we[in]}); Commenters are more likely to refer to in-group using the most inclusive term at higher WPs, when referring to the in-group at all. The tendency for commenters to use the first-person singular pronoun (which we categorized as the most inclusive referent from our taxonomy described in \textsection~\ref{subsec:mereology}) more often at lower WPs can be further explained by fans strategizing what their team \emph{should do}, to increase their WP (and thus come back in the game):

\begin{table}[t]
    \centering
    \begin{tabular}{lrr}
        \toprule
        \textbf{Feature} & \textbf{Slope} & \textbf{R-squared} \\
         & ($\times$10\textsuperscript{-4}) & \\\midrule
        Any reference([any]) & -19.3 & 0.72 \\ 
        No reference(none) & 2.4 & 0.65 \\\midrule
        In-group([in]) & -2.8 & 0.31 \\ 
        we[in] & -2 & 0.61 \\  
        Out-group([out]) & 2.5 & 0.56 \\
        they[in] & -0.3 & 0.15 \\
        they[out] & 0.4 & 0.25 \\ \bottomrule
    \end{tabular}
    \caption{Table of slopes of feature of interest against increasing WP, alongside the r-squared showing how much of the variance is explained by the linear regression fit. The slopes for Any and no reference are calculated with frequencies normalized by total number of referents in a WP window. All other slopes are measured with frequencies normalized with only those comments that have references in that WP window.}
    \label{tab:slopes}
\end{table}

\ex. \a. If \textbf{we} get a stop here and a touchdown on the next drive it's a ballgame let's fucking go .	
     \b. We need a rebuild of the players and coaching staff

These findings add to the subtle ways we perpetuate bias in our linguistic behavior, especially towards \textbf{in-group protection}~\citep{maass_linguistic_1999}. While commenters are more than willing to criticize the in-group across WP, the self-protective instinct is evident in the way they choose to refer to the in-group using \emph{we/us} forms more often when losing, the reduced tendency to refer to the in-group using \emph{they/them}, or to not refer to the in-group at all when winning. Thus, the form of referring expression  commenters use to refer to the in/out-group represents just as subtle a bias as the predicate form variation hypothesized by the LIB.

\section{Conclusion}
\label{sec:conclusion}
We enhance our understanding of intergroup bias by building a parallel corpus of sports comments grounded in win probabilities from live games. Annotation experiments show that modeling the bias as a tagging problem over referring words can reveal unobserved variations, as well as make it amenable to large-scale modeling. Through few-shot and finetuning experiments, we find that LLMs can out-match human baseline performance at this task, but struggle to reason over win probabilities, or use it meaningfully towards tagging. Tagging a large sample of our dataset reveals linear trends between various referring expressions and WP, showing that intergroup bias can manifest in commenter's choice of who to refer to when commenting on a game and how. Careful data curation and understanding, combined with focused usage of LLMs as statistical information processing tools can thus uncover linguistic variations in social language use online at scale. In future work we plan to exploit the parallel nature of our corpus further to understand team differences in language variation, as well as how WP can be effectively incorporated into a model of social meaning. 

\section*{Limitations}
\label{sec:limits}

Our work expands the study of intergroup bias in language by focusing on natural language use in online conversations on the Reddit platform. Further, our focus on grounding the utterances lead us to focus on sports talk, specifically conversations around NFL games. Biases in demographics of users on Reddit, or demographics of NFL fans are thus inherent in our data and analysis. Future work needs to study the prevalence of our findings in other sports with similar statistics that enables efficient grounding of utterances, as well as in more general speech.

We identify that both few-shot performance by GPT-4o and finetuned performance by Llama-3 are close to, or out-perform the human ceiling performance. Human ceiling performance is simply the average accuracy of crowd annotators against expert annotators. As we note in the paper, this is a difficult and inherently subjective task. Our results do not mean that models (finetuned or not) have a better understanding of what constitutes intergroup references, nor that they are more aligned with the task. Llama-3 was trained on the training split of the expert annotated gold data-set. While GPT-4o was exposed to the same set of examples as human annotators, it is a very large (possibly a mixture of trillions of parameters) model that contains a multitude of statistical associations that aids in instruction following.

\section*{Ethics}
\label{sec:ethics}

We downloaded comments from Reddit threads using the official Reddit API, and will disseminate our data in accordance with the Reddit terms of service. We will only release the comment and post ids for the raw data, and usernames will be anonymized. We will release the annotated data in full with the same precautions. We have censored some of the profanity in the comments when used as examples in this paper, since our focus isn't on abusive/negative language exclusively. All created artifacts from this work (code, annotated data) will be released under the MIT License. Crowd-sourced annotations were collected from three undergraduates employed by one of the  authors for \$15 an hour.

\section*{Acknowledgments}

This research is partially supported by Good Systems\footnote{\url{https://goodsystems.utexas.edu}}, a UT Austin Grand Challenge to develop responsible AI technologies, and NSF grants IIS-2145479, IIS-2107524. We thank our annotators Akhila Gunturu, Kathryn Kazanas and Melanie Quintero for their efforts, feedback and highlighting interesting examples from our dataset. We thank Malihe Alikhani and Kanishka Misra for their valuable feedback of this work. We would also like to thank Texas Advanced Computing Center (TACC) for providing computational resources used to run some of the experiments in this paper.

\bibliography{references}
\bibliographystyle{acl_natbib}

\appendix
\section{Data}
\label{app:data}

\begin{figure}[H]
    \centering
    \includegraphics[width=\linewidth]{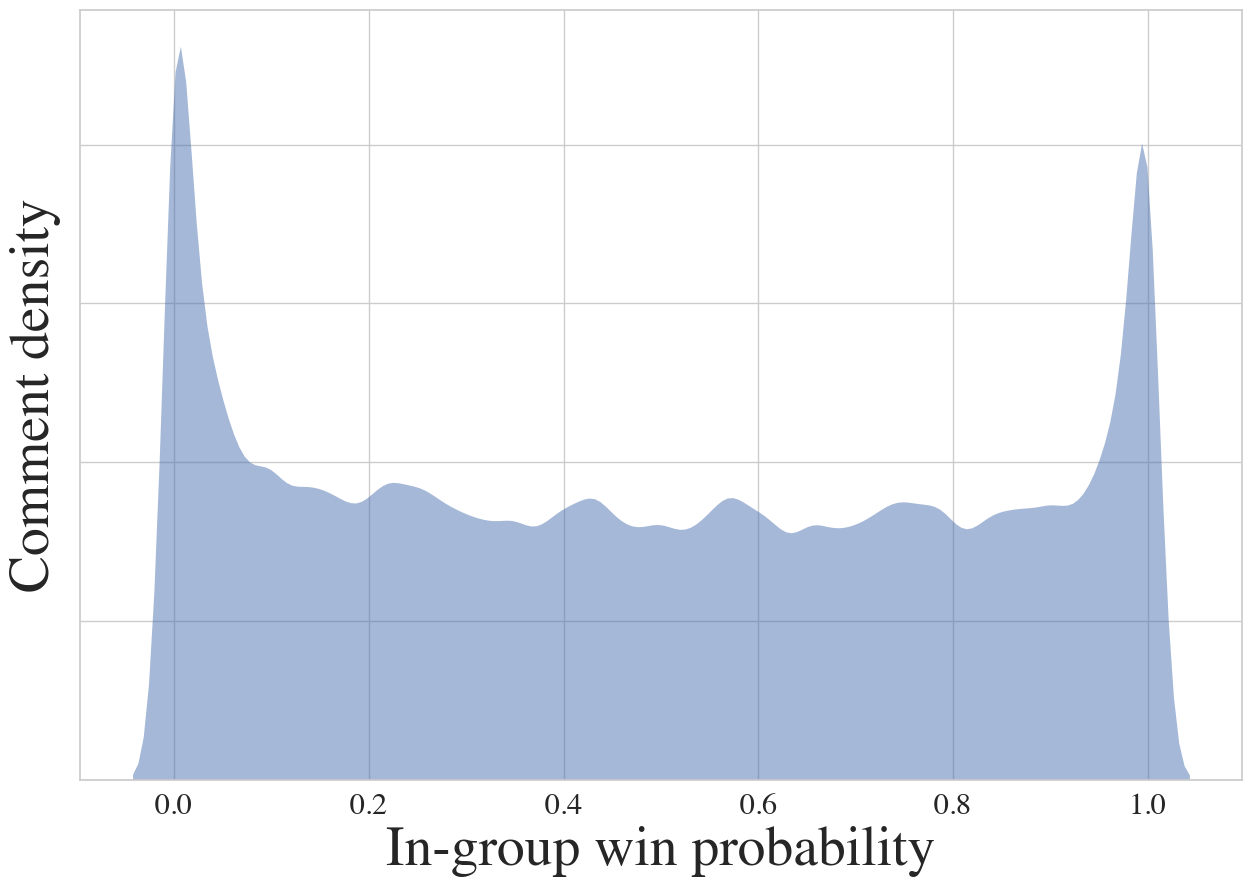}
    \caption{Comment density against WP.}
    \label{fig:density}
    \vspace{-\baselineskip}
\end{figure}

\section{Annotation}
\label{app:ann}

\paragraph{Protocol} Annotators were given the following instructions:

\begin{enumerate}
    \item All comments are from game threads corresponding to specific NFL games between two teams. You will be given the source of the comment --- this is the team the writer of the comment supports, the opponent in that game, and the live score at the time of making the comment.
    \item Highlight any words and phrases that refer to individuals (people, teams, sub-groups within the team, organizations).
    \item If the reference is to the same group as the source subreddit of the comment, tag this highlight as \textbf{in-group} ({\scshape [in]}).
    \item If the reference is towards the opponent in this specific game for which the comment is written, tag this highlight as \textbf{out-group} ({\scshape [out]}).
    \item If the reference is towards any other team in the NFL apart from the two teams involved in this game, tag this highlight as \textbf{other} ({\scshape [other]}).
    \item Some comments will not have an obvious reference to an in-group/out-group/other entity. Leave these comments un-annotated. If you’re unsure of an annotation, you can indicate your confidence, but only use the confidence scale if you’re not very confident with your annotation. I will take an empty confidence annotation as full confidence.
    \item Do not annotate a {\scshape [sent]} token if there is a word in the sentence that can be annotated with the same label.
\end{enumerate}

They were also given the following examples. Models were finetuned with the following as few-shot examples --- they were provided WP over live score for a more holistic representation of the game, and explanations were modified depending on whether WP was provided or not.

\begin{prompt}[title={\thetcbcounter{} Example 1},label=example:1]
{\scshape comment}: {\scshape [sent]} Defense getting absolutely bullied by a dude that looks like he sells solar panels

{\scshape parent comment}: None

{\scshape in-group}: Jets

{\scshape out-group}: Bears

{\scshape live score}: Jets 7 - 3 Bears

{\scshape target}: {\scshape [sent]} {\scshape [in]} getting absolutely bullied by {\scshape [out]} that looks like {\scshape [out]} sells solar panels .

{\scshape explanation}: The commenter is probably talking about the in-group, since 'Defense' is said without qualification, and the description of the offensive player is disparaging ('he sells solar panels'). 'Defense' should be tagged {\scshape [in]} since it refers to in-group, and 'a dude' and 'he' should be tagged {\scshape [out]} since it refers to an out-group offensive player.

\end{prompt}

\begin{prompt}[title={\thetcbcounter{} Example 2},label=example:2]
{\scshape comment}: {\scshape [sent]} Hasn’t really been him . {\scshape [sent]} Receivers have been missing a lot of easy catches.

{\scshape parent comment}: Dont know maybe Tua is choking, all the pressure

{\scshape in-group}: Dolphins

{\scshape out-group}: Chargers

{\scshape live score}: Dolphins 0 - Chargers 0

{\scshape target}: {\scshape [sent]} Hasn’t really been {\scshape [in]} . {\scshape [sent]} {\scshape [in]} have been missing a lot of easy catches .

{\scshape explanation}: The second sentence is complaining about the receivers missing a lot of catches, thus absolving another player of some blame, which is something fans would only do for the in-group team they support. Thus 'him' in first sentence, and 'Receivers' in second sentence should be tagged with {\scshape [in]}.
\end{prompt}

\begin{prompt}[title={\thetcbcounter{} Example 3},label=example:3]
{\scshape comment}: {\scshape [sent]} Cards and rams are gonna be in the post-season regardless, so I don't really care about them losing unless they play us.

{\scshape parent comment}: Who do y'all want to lose this afternoon for Cards/Seahawks game?

{\scshape in-group}: 49ers

{\scshape out-group}: Jaguars

{\scshape live score}: 49ers 30 - 10 Jaguars

{\scshape target}: {\scshape [sent]} {\scshape [other]} and {\scshape [other]} are gonna be in the post-season regardless, so I don't really care about {\scshape [other]} losing unless they play {\scshape [in]}.

{\scshape explanation}: The game is between the 49ers and Jaguars, while the words 'Cards' and 'rams' refers to other teams in the NFL. Thus they should be tagged {\scshape [other]} since they are neither in-group nor out-group, as should the word 'them'. 'us' should be tagged {\scshape [in]} since it refers to the in-group team the player supports.
\end{prompt}

\begin{prompt}[title={\thetcbcounter{} Example 4},label=example:4]
{\scshape comment}: {\scshape [sent]} How are we this shit on defense

{\scshape parent comment}: None

{\scshape in-group}: Steelers

{\scshape out-group}: Eagles

{\scshape live score}: Steelers 7 - 21 Eagles

{\scshape target}: {\scshape [sent]} How are {\scshape [in]} this shit on defense

{\scshape explanation}: 'we' here, and almost always, refers to the in-group since they don't like their team's defense, which is reflected in the score. 'we' should therefore be tagged with {\scshape [in]} since it refers to in-group.
\end{prompt}

\begin{prompt}[title={\thetcbcounter{} Example 5},label=example:5]
{\scshape comment}: {\scshape [sent]} The chiefs got straight fucked with that Herbert INT getting called dead . {\scshape [sent]} Suck it , KC !	

{\scshape parent comment}: None

{\scshape in-group}: Chargers

{\scshape out-group}: Chiefs

{\scshape live score}: Chargers 28 - 28 Chiefs

{\scshape target}: {\scshape [sent]} {\scshape [out]} got straight fucked with that {\scshape [in]} INT getting called dead . {\scshape [sent]} Suck it , {\scshape [out]} !

{\scshape explanation}: This is a game between the Chiefs and the Chargers, and the commenter is a supporter of the Chiefs, so 'the chiefs' in the first sentence and 'KC' in the second sentence should be tagged {\scshape [out]}. Herbert is a player for the Chargers, and should be tagged with {\scshape [in]} since he is a member of the in-group with respect to the commenter.
\end{prompt}

\begin{prompt}[title={\thetcbcounter{} Example 6},label=example:6]
{\scshape comment}: {\scshape [sent]} Need points but 7 would be HUGE momentum

{\scshape parent comment}: None

{\scshape in-group}: Bengals

{\scshape out-group}: Chiefs

{\scshape live score}: Bengals 3 - 13 Chiefs

{\scshape target}: {\scshape [in]} Need points but 7 would be HUGE momentum

{\scshape explanation}: The in-group team is losing currently as the score shows, so this comment is implicitly about the in-group needing points to gain momentum. Thus '{\scshape [sent]}' should be tagged with '{\scshape [in]}' since there is no explicit word/phrase that refers to the in-group, but the comment is referring to the in-group implicitly.
\end{prompt}

\section{Prompts}
\label{app:prompts}

Below is the prompt provided to both GPT-4o and Llama-3. Examples are the same as the ones provided to human annotators, listed in the previous section. The following prompt does not use win probabilities; The prompts which do use WP are the same as below, except they include a definition of WP as `the probability of the in-group winning the game at the time of the comment - if the win probability is high, the in-group team is probably doing well and going to win.' in the prompt text.

\begin{prompt}[title={\thetcbcounter{} Prompt},label=prompt:1]
Tag references to entities as in-group ({\scshape [in]}), out-group ({\scshape [out]}) or other ({\scshape [other]}) in live, online sports comments during NFL games. The input is the comment, the parent comment (if the comment is a reply, else it will be 'None'), the in-group team the commenter supports and the out-group opponent team during that game. Using knowledge of American football and contextual language understanding, identify words and phrases denoting entities (players, teams, city names, sub-groups within the team) that refer to the in-group ({\scshape [in]} - team the commenter supports), out-group ({\scshape [out]} - the opponent) or other teams ({\scshape [other]} - some other team in the NFL that is not the in-group or the opponent), with respect to the commenter. Return the list of words/phrases that are to be tagged ({\scshape ref\_expressions}), an {\scshape explanation} reasoning over why these words and phrases in {\scshape comment} should be tagged and with what tag, and the {\scshape target} comment itself with relevant words/phrases replaced with the respective tags ({\scshape [in]}, {\scshape [out]} or {\scshape [other]}) in your final output.

\hfill

Each sentence in a comment is separated by a {\scshape [sent]} token. Sometimes a sentence in the comment will be about the in/out/other group but not have an explicit word/phrase that refers to the group; In such cases, tag the {\scshape [sent]} token for that sentence with the corresponding tag label.

\hfill

Here are 6 examples, with {\scshape ref\_expressions} being the list of words/phrases to be tagged from {\scshape comment}, {\scshape explanation} being a reasonable reason for why these words/phrases should be tagged with appropriate tags, and {\scshape target} being the correct tagged output for {\scshape comment}.

\hfill

[Examples 1-6 follow]

\hfill

Some comments will have no explicit or implicit reference to the in-group, out-group, or other, or it could be extremely hard to disambiguate any references based on given information. In such cases, return Target as a copy of Comment, justify this with the Explanation, "No explicit or implicit references to tag.", and return [] for {\scshape ref\_expressions}. Here is an example:

\hfill

{\scshape comment}: {\scshape [sent]} I thought so. {\scshape [sent]} Wish I could say the same ;)

{\scshape parent comment}: Great input

{\scshape in-group}: Jaguars

{\scshape out-group}: Titans

{\scshape ref\_expressions}: []

{\scshape explanation}: No explicit or implicit references to tag.

{\scshape target}: {\scshape [sent]} I thought so. {\scshape [sent]} Wish I could say the same ;)

\hfill

Now tag only the relevant words/phrases in the following comment as either in-group ({\scshape [in]}), out-group ({\scshape [out]}), or other ({\scshape [other]}), if any. First return the list of words to be tagged, then explain your reasoning as to why these words/phrases should be tagged from {\scshape comment} and with which tags, and finally return the tagged comment in that order.

\end{prompt}

The explanations in the prompt with WP are similar to the explanations provided previously. Here is the explanation for Example 1 from the previous section using win probabilities instead of the live score:

\begin{prompt}[title={\thetcbcounter{} Explanation-WP}]
The commenter is probably talking about the in-group, since 'Defense' is said without qualification, and the description of the offensive player is disparaging ('he sells solar panels'). This is in spite of the win probability being relatively high for the in-group - sometimes commenters choose to focus on immediate plays rather than the overall state of the game, so perhaps this comment was in response to a bad showing by the defense. 'Defense' should be tagged {\scshape [in]} since it refers to in-group, and 'a dude' and 'he' should be tagged {\scshape [out]} since it refers to an out-group offensive player.
\end{prompt}

Here is an example model input after the prompt:

\begin{prompt}[title={\thetcbcounter{} Model Input}]
{\scshape comment}: {\scshape [sent]} Defense getting absolutely bullied by a dude that looks like he sells solar panels.

{\scshape parent comment}: None

{\scshape in-group}: Jets

{\scshape out-group}: Bears

{\scshape win probability}: 71.5\%

{\scshape ref\_expressions}:
\end{prompt}

and here is the model's expected output:

\begin{prompt}[title={\thetcbcounter{} Model Output}]
['Defense', 'a dude', 'he']

{\scshape explanation}: The commenter is probably talking about the in-group,since 'Defense' is said without qualification, and the description of the offensive player is disparaging ('he sells solar panels'). 'Defense' should be tagged {\scshape [in]} since it refers to in-group, and 'a dude' and 'he' should be tagged {\scshape [out]} since it refers to an out-group offensive player.

{\scshape target}: {\scshape [sent]} {\scshape [in]} getting absolutely bullied by {\scshape [out]} that looks like {\scshape [out]} sells solar panels .
\end{prompt}

Explanations from GPT-4o were generated using the same prompt format --- the model is tasked with generating {\scshape explanation} instead of {\scshape target}. The same examples were used, with {\scshape explanation} at the end rather than {\scshape target}. GPT-4o was tasked to generate explanations of maximum length 256 tokens with seed of 1.

\paragraph{Linguistic WP} In the linguistic WP condition, we replace the percentage WP value with a text string like below:

\begin{itemize}
    \item 0--25: \emph{Team A} is very likely to lose.
    \item 25--45: \emph{Team A} is likely to lose.
    \item 45--55: Both teams are equally likely to win.
    \item 55--75: \emph{Team B} is likely to win.
    \item 75--100: \emph{Team B} is very likely to win.
\end{itemize}

\section{Modeling implementation}
\label{app:implement}

\paragraph{GPT-4o} All few-shot experiments were run with \texttt{gpt-4o-2024-05-13}. Temperature was set to 1 if temperature scaling wasn't used, else it is dynamically set to $sin(\pi\times WP)$.

\paragraph{Llama-3-8B} We finetuned the base \texttt{llama-3-8b} model from Meta's Huggingface model space\footnote{\url{https://huggingface.co/meta-llama/Meta-Llama-3-8B}}. We used the \texttt{Axolotl}\footnote{\url{https://github.com/OpenAccess-AI-Collective/axolotl}} framework for all fine-tuning experiments with the following hyper-parameter settings:

\begin{itemize}
    \item batch size of 4 for training and inference.
    \item sample packing and padding to sequence length were enabled, with a max sequence length of 2560. None of our inputs exceeded this limit.
    \item Cosine learning rate scheduler with warmup of 10 steps, learning rate set to $1e-5$, weight decay of 0.1, and a minimum learning rate ratio of 0.1
    \item Maximum of 2 train epochs with early stopping, and patience set to 3.
    \item The model is evaluated and saved every 59 steps for a maximum of 595 steps.
    \item Flash attention and gradient checkpointing were enabled.
\end{itemize}

All finetuning experiments were done on 2 Nvidia A40 GPUs, and each fine-tuning run took approximately 1.5 hours.

\end{document}